\documentclass[12pt]{article}

\usepackage{amssymb,amsmath,amsfonts,latexsym,graphicx}
\usepackage[numbers]{natbib}
\usepackage{xurl}
\usepackage{xcolor}
\usepackage{hyperref}
\hypersetup{
    colorlinks=true,   
    linkcolor=blue,    
    citecolor=blue,    
    urlcolor=blue      
}

\begin{document}

\begin{center}
\Large \bf Capsule Endoscopy Multi-classification via Gated Attention and Wavelet Transformations \rm

\vspace{1cm}


\large Lakshmi Srinivas Panchananam $^a$, \large  Praveen Kumar Chandaliya $^b$, \\ \large Kishor Upla $^c$, Kiran Raja $^d$

\vspace{0.5cm}

\normalsize


$^a$ National Institute of Technology Puducherry, India

$^b \, ^c$ Sardar Vallabhbhai National Institute of Technology Surat, India

$^d$ Norwegian University of Science and Technology, Norway

\vspace{5mm}

{\tt $^a$10srinu24@gmail.com, $^a$cs22b1040@nitpy.ac.in, $^b$pkc@aid.svnit.ac.in,\\ $^c$kpu@eced.svnit.ac.in, $^d$kiran.raja@ntnu.no}

\vspace{1cm}

\end{center}

\abstract{Abnormalities in the gastrointestinal tract significantly influence the patient's health and require a timely diagnosis for effective treatment. With such consideration, an effective automatic classification of these abnormalities from a video capsule endoscopy (VCE) frame is crucial for improvement in diagnostic workflows. 

The work presents the process of developing and evaluating a novel model designed to classify gastrointestinal anomalies from a VCE video frame. Integration of Omni Dimensional Gated Attention (OGA) mechanism and Wavelet transformation techniques into the model's architecture allowed the model to focus on the most critical areas in the endoscopy images, reducing noise and irrelevant features. This is particularly advantageous in capsule endoscopy, where images often contain a high degree of variability in texture and color. Wavelet transformations contributed by efficiently capturing spatial and frequency-domain information, improving feature extraction, especially for detecting subtle features from the VCE frames. Furthermore, the features extracted from the Stationary Wavelet Transform and Discrete Wavelet Transform are concatenated channel-wise to capture multiscale features, which are essential for detecting polyps, ulcerations, and bleeding. This approach improves classification accuracy on imbalanced capsule endoscopy datasets. The proposed model achieved $92.76\%$ and $91.19\%$ as training and validation accuracies respectively. At the same time, Training and Validation losses are $0.2057$ and $0.2700$. The proposed model achieved a Balanced Accuracy of $94.81\%$, AUC of $87.49\%$, F1-score of $91.11\%$, precision of $91.17\%$, recall of $91.19\%$ and specificity of $98.44\%$. Additionally, the model's performance is benchmarked against two base models, VGG16 and ResNet50, demonstrating its enhanced ability to identify and classify a range of gastrointestinal abnormalities accurately. This work secured 27th rank in the Capsule Vision 2024 Challenge. The implementation and additional resources can be found at \url{https://github.com/09Srinivas2005/Capsule-Endoscopy-Multi-classification-via-Gated-Attention-and-Wavelet-Transformations.git}.

\section{Introduction}\label{sec1}

\begin{figure}[h]
    \centering
    \includegraphics[width=0.80\linewidth]{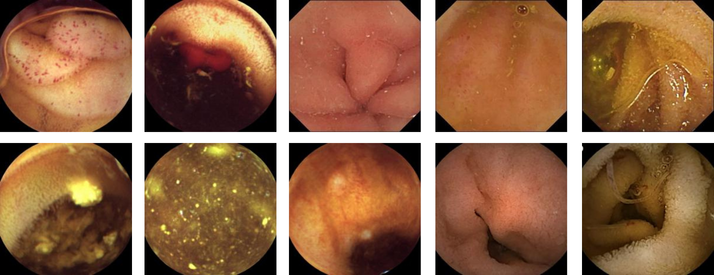}
    \caption{Sample Images from Dataset \cite{Handa2024TrainingDataset}}
    \label{figure:SampleImages}
\end{figure}

\begin{figure}[h]
    \centering
    \includegraphics[width=0.80\linewidth]{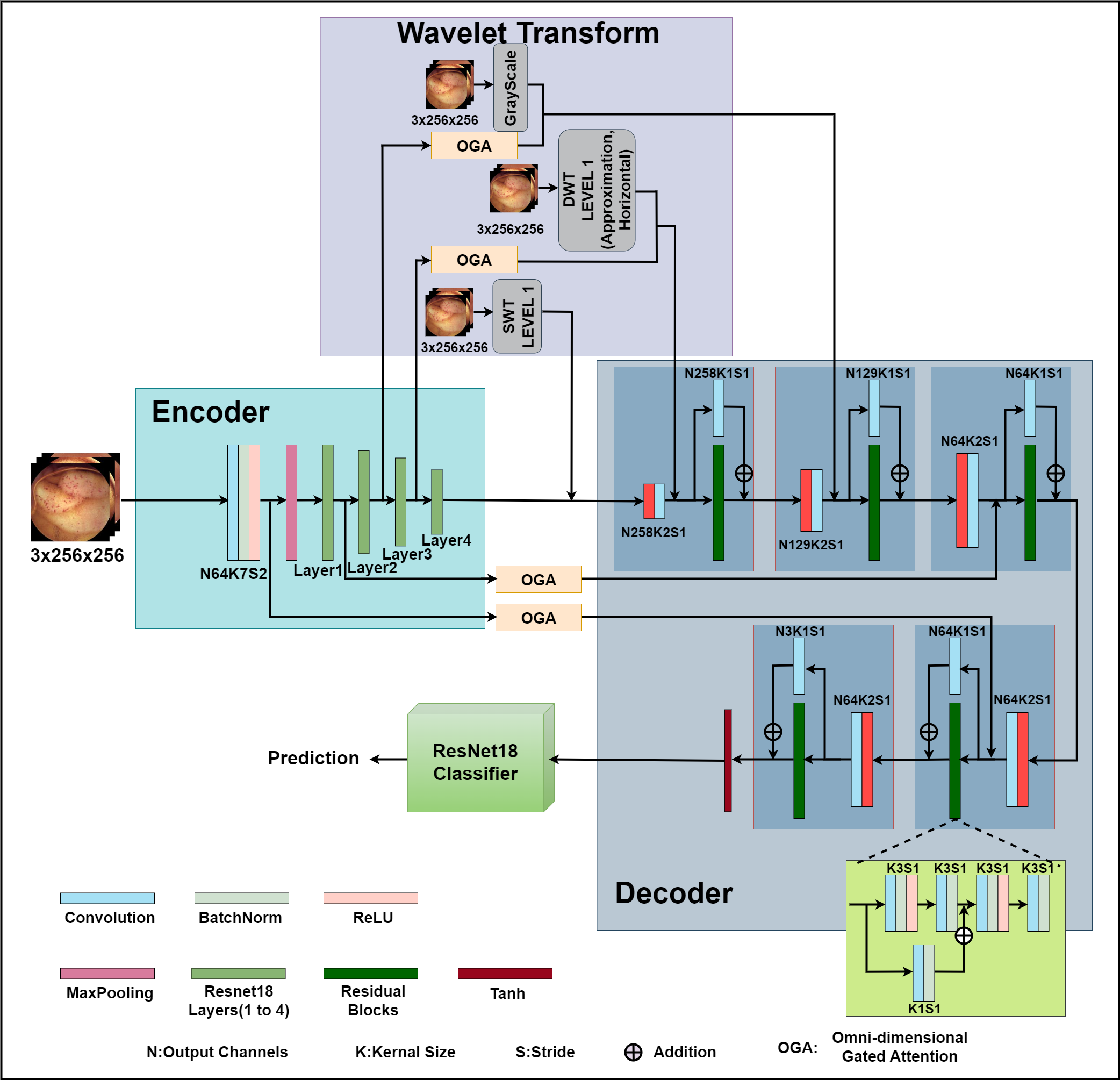}
    \caption{Proposed model architecture.}
    \label{figure:NetworkArchitecture}
\end{figure}

\begin{figure}[h]
    \centering
    \includegraphics[width=0.80\linewidth]{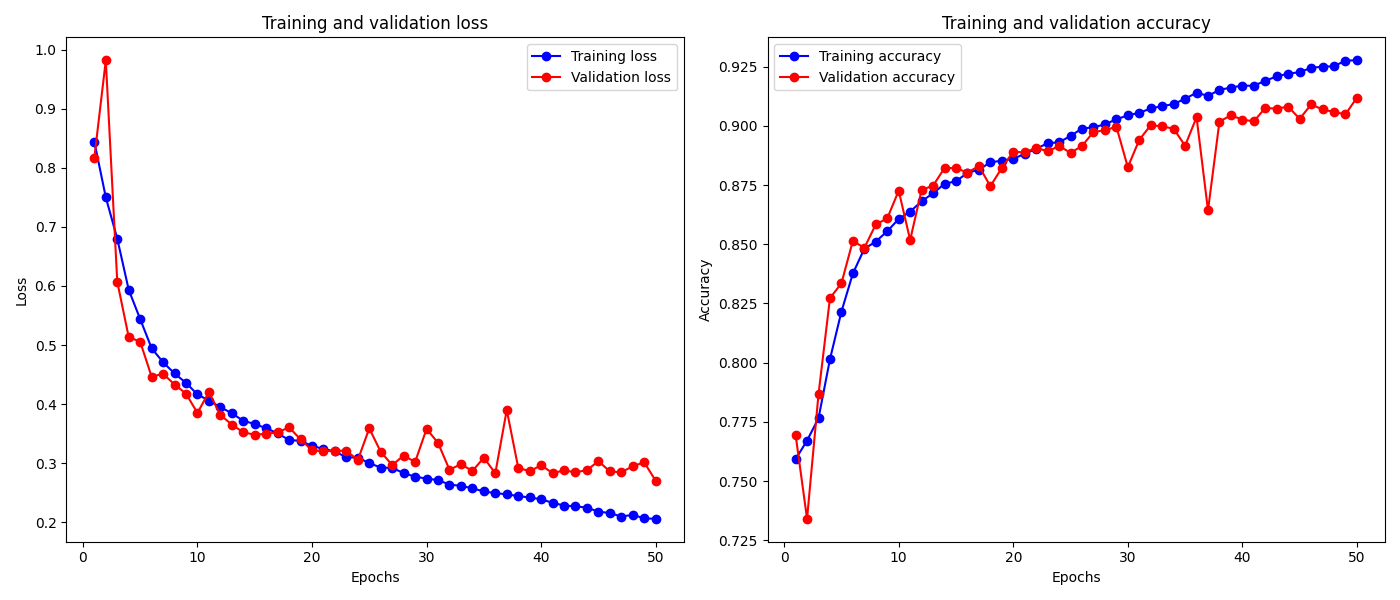}
    \caption{Training and Validation loss and accuracy.}
    \label{figure:TrainValidationPlot}
\end{figure}

Gastrointestinal abnormalities like bleeding, polyps, and ulcers may lead to serious medical complications if not diagnosed and treated promptly. A breakthrough in the approach taken in discovering these abnormalities is with the advent of Video Capsule Endoscopy \cite{Bresci2005}, which allows for a non-invasive imaging of the gastrointestinal tract. It requires tremendous human time, is prone to human error, and relies heavily on clinical expertise for the analysis of a very large number of frames generated by a VCE. The recent emergence of deep learning models as powerful tools for the automation of disease detection could thus improve the accuracy of diagnosis, reduce workload, and ensure timely interventions. These models will automatically extract relevant features and patterns from medical images, hence offering more accurate identification of abnormalities.

In this work, we introduce an innovative deep-learning model for the automated classification of gastrointestinal abnormalities in VCE video frames. The novelty of our model arises from combining the Omni Dimensional Gated Attention and wavelet transformation techniques to enhance feature extraction. OGA dynamically generates spatial, channel, filter, and kernel-wise attentions to allow the model to focus on relevant features in each dimension. Moreover, the model captures multi-scale information by using SWT and DWT features, which are crucial for accurately distinguishing between different types of abnormalities.

Our model demonstrates superior performance in terms of both accuracy and interpretability. By introducing OGA to our model, it dynamically adjusts the attention weights across dimensions. This improves feature learning and hence enhances classification accuracy. In addition, the use of wavelet transforms proves to be a robust mechanism for capturing multi-scale patterns. In contrast, previous models often tend to ignore such multi-scale patterns. We do a comparison between VGG16 and ResNet50. This comparison reflects the efficiency of our approach in the improvement of the given AUC, accuracy, and specificity.

\section{Methods}\label{sec2}

The network architecture, as shown in the figure \ref{figure:NetworkArchitecture}, works in four stages. (i) Multi-level feature extraction by the encoder, (ii) Integrating Wavelet Transformed features \cite{AlSuwaidi_2018}, (iii) Reconstructing Images by the decoder, (iv) Classification.
The encoder performs hierarchical feature extraction to generate detailed and robust intermediate representations, which are then passed through Omni Dimensional Gated Attention (OGA). Inspired by Phutke et al.\cite{OGA}, the omni-dimensional gated attention (OGA) mechanism is integrated, to dynamically generates spatial, channel, filter, and kernel-wise attentions for encoder features, allowing the network to capture weighted features across all dimensions for better feature learning and decoding on an imbalanced dataset. This feature set is further enhanced by concatenating those features channel-wise with Stationary Wavelet Transform Features and Discrete Wavelet Transform features. In the following stage, the decoder utilizes this enhanced feature set to reconstruct the images, passed to the classifier for the final video capsule endoscopy classification.

\section{Results}\label{sec3}

This section presents a brief description of the experimental setup and the performance evaluation of the model trained on Dataset \cite{Handa2024TrainingDataset} and assessed on Dataset \cite{Handa2024TestingDataset}.
\subsection{Experimental Setup}
   For the training process, the images were preprocessed by resizing the images to dimensions $256 \times 256$ and normalizing them before feeding into the proposed model. The training was performed with a batch size of $16$, with four workers for data loading while training and two workers during validation. The model was trained for $50$ epochs with a learning rate of $0.001$. Cross Entropy Loss was employed as the loss function, and the Adam optimizer with L2 regularization (weight decay of 1e-5) was used to mitigate overfitting. The model was implemented using PyTorch.

    All experiments were conducted on an Artificial Intelligence server running Ubuntu 22.04.2 LTS. The server is equipped with NVIDIA A100-SXM4-40GB GPUs, featuring a total of $8$ GPUs, each with $40$ GB of memory. $535.183.01$ and $12.2$ are the drivers and the CUDA versions of the GPUs. The server utilizes AMD EPYC 7742 64-core Processors. The CPU core count of its $x86\_64$ architecture is $128$ and its frequency range is between $1.5$ GHz and $2.25$ GHz. Furthermore, the server has $1.0$ TiB of RAM installed. 

\subsection{Quantitative Results}
The proposed model is trained and evaluated on dataset \cite{Handa2024TrainingDataset} and dataset \cite{Handa2024TestingDataset} respectively. Figure \ref{figure:TrainValidationPlot} shows the training and validation plot. Additionally, its performance is compared to the baseline models (VGG16, ResNet50) using metrics such as AUC, Balanced Accuracy, Validation Accuracy, F1-score, Precision, Recall, and Specificity as shown in the Table \ref{tab:results}. The proposed model achieved an average accuracy of $90.42\%$, outperforming VGG16 ($69.06\%$) and ResNet50 ($76.02\%$). This improvement underscores the effectiveness of the gated attention mechanism and wavelet transformation in distinguishing between different gastrointestinal conditions. The model attained a precision of $89.73\%$, recall of $90.42\%$, and F1-score of $89.82\%$ across multiple classes of gastrointestinal abnormalities. These metrics reflect a balanced performance, ensuring both high sensitivity (recall) and precision in identifying abnormalities, which is critical in minimizing false positives and false negatives in clinical applications.
\label{subsec1}


\begin{table*}[h]
\centering
\resizebox{\textwidth}{!}{
\begin{tabular}{cccccccc}
\hline

\textbf{Method} & \textbf{\begin{tabular}[c]{@{}c@{}}Mean \\ AUC\end{tabular}} & \textbf{\begin{tabular}[c]{@{}c@{}}Balanced \\ Accuracy\end{tabular}} & \textbf{\begin{tabular}[c]{@{}c@{}}Avg. \\ Accuracy\end{tabular}} & \textbf{\begin{tabular}[c]{@{}c@{}}Avg. \\ F1-score\end{tabular}} & \textbf{\begin{tabular}[c]{@{}c@{}}Avg. \\ Precision\end{tabular}} & \textbf{\begin{tabular}[c]{@{}c@{}}Avg. \\ Recall\end{tabular}} & \textbf{\begin{tabular}[c]{@{}c@{}}Avg. \\ Specificity\end{tabular}} \\ \hline
\textbf{VGG16} & 91.61 & 56.84 & 69.06 & 48.44 & 52.46 & 54.30 & 96.97 \\ 
\textbf{ResNet50} & 87.10 & - & 76.02 & 76.0 & 78.0 & 76.0 & - \\ \hline
\textbf{Our Model} & 87.49 & \textbf{94.81} & \textbf{91.19} & \textbf{91.11} & \textbf{91.17} & \textbf{91.19} & \textbf{98.44} \\ \hline
\end{tabular}}
\caption{Results on the validation dataset were compared to the baseline methods reported by the organizing team of Capsule Vision 2024.}
\label{tab:results}
\end{table*}

\section{Discussion}\label{sec4}
Unlike traditional CNN-based methods such as VGG16 and ResNet50, which rely solely on spatial features, the proposed approach leverages dual-domain analysis, enabling the model to detect features across spatial and frequency domains. The gated attention mechanism plays a crucial role in dynamically allocating attention, enhancing the identification of disease-relevant regions in an imbalanced dataset. Additionally, the wavelet transformation’s ability to decompose images into multiple frequency bands provides richer information for distinguishing pathological from non-pathological regions, as demonstrated in Table \ref{tab:results}.

\section{Conclusion}\label{sec5}

We propose a novel architecture that integrates Omni Dimensional Gated Attention (OGA) \cite{OGA} and wavelet transformation \cite{AlSuwaidi_2018} techniques into a ResNet-based encoder-decoder framework for the automatic classification of gastrointestinal abnormalities. The encoder leverages a modified ResNet18 backbone \cite{Xie} to extract intricate multi-level feature representations from the input frames. This feature set is further enhanced by the incorporation of OGA and Wavelet transformation, which allows the model to capture fine and subtle information from an imbalanced dataset. The decoder then reconstructs the images from this enhanced feature set, which are then processed by a classifier to predict the final class of gastrointestinal anomalies.

Results of performance analysis by the proposed model indicate its strength in the efficient classification of different types of gastrointestinal abnormalities with an AUC of $84.83\%$ and balanced accuracy of $94.05\%$. Compared to baseline models VGG16 and ResNet50, our model outperforms the existing approaches in accurately classifying and distinguishing between different types of abnormalities. Consistency across the different metrics shows the model's versatility and robustness, making it a powerful tool in the classification of gastrointestinal abnormalities in Video Capsule Endoscopy (VCE) frames.

\section{Acknowledgments}\label{sec6}
As participants in the Capsule Vision 2024 Challenge, we fully comply with the competition's rules as outlined in \cite{handa2024capsule}. Our AI model development is based exclusively on the datasets provided in the official release in \cite{Handa2024TrainingDataset}.

\bibliographystyle{unsrtnat}
\bibliography{sample}

\end{document}